\def\year{2021}\relax
\documentclass[letterpaper]{article} 
\usepackage{aaai21}  
\usepackage{times}  
\usepackage{helvet} 
\usepackage{courier}  
\usepackage[hyphens]{url}  
\usepackage{graphicx} 
\usepackage{natbib}  
\usepackage{caption} 
\usepackage[utf8]{inputenc} 
\usepackage[T1]{fontenc}    
\usepackage{url}            
\usepackage{booktabs}       
\usepackage{amsfonts}       
\usepackage{nicefrac}       
\usepackage{microtype}      

\usepackage{amsmath,amssymb,amsthm,bbm}
\usepackage{mathtools}
\usepackage{microtype}
\usepackage{booktabs}
\usepackage{multirow}
\usepackage{makecell}
\usepackage{subcaption}
\usepackage{color}
\usepackage{xcolor}

\newcommand{\ie}{\textit{i}.\textit{e}.}
\newcommand{\eg}{\textit{e}.\textit{g}.}

\DeclarePairedDelimiter\norm{\lVert}{\rVert}

\newcommand{\tf}{\mathtt{tf}}
\newcommand{\idf}{\mathtt{idf}}

\newcommand{\stdv}[1]{\scriptsize$\pm$#1}

\newcommand{\cell}[3]{{
	\makecell{#1\stdv{#2}\\\rel{#3}}
}}

\title{MASKER: Masked Keyword Regularization for Reliable Text Classification}
\author{
Seung Jun Moon\thanks{Equal contribution}$^1$, Sangwoo Mo$^{*1}$, Kimin Lee$^{2}\thanks{Work was done while the author was at KAIST}$, Jaeho Lee$^1$, Jinwoo Shin$^1$\\
}
\affiliations{
$^1$Korea Advanced Institute of Science and Technology, South Korea\\
$^2$University of California, Berkeley, USA\\
\texttt{\{june1212,swmo,jaeho-lee,jinwoos\}@kaist.ac.kr} \quad \texttt{kiminlee@berkeley.edu}\\
}

\begin{document}
\maketitle

\begin{abstract}
Pre-trained language models have achieved state-of-the-art accuracies on various text classification tasks, \eg, sentiment analysis, natural language inference, and semantic textual similarity. However, the \textit{reliability} of the fine-tuned text classifiers is an often underlooked performance criterion. For instance, one may desire a model that can detect out-of-distribution (OOD) samples (drawn far from training distribution) or be robust against domain shifts. We claim that one central obstacle to the reliability is the over-reliance of the model on a limited number of keywords, instead of looking at the whole context. In particular, we find that (a) OOD samples often contain in-distribution keywords, while (b) cross-domain samples may not always contain keywords; over-relying on the keywords can be problematic for both cases. In light of this observation, we propose a simple yet effective fine-tuning method, coined masked keyword regularization (MASKER), that facilitates context-based prediction. MASKER regularizes the model to reconstruct the keywords from the rest of the words and make low-confidence predictions without enough context. When applied to various pre-trained language models (\eg, BERT, RoBERTa, and ALBERT), we demonstrate that MASKER improves OOD detection and cross-domain generalization without degrading classification accuracy. Code is available at \url{https://github.com/alinlab/MASKER}.
\end{abstract}

\section{Introduction}
\label{sec:intro}

Text classification \citep{aggarwal2012survey} is a classic yet challenging problem in natural language processing (NLP), having a broad range of applications, including sentiment analysis \citep{bakshi2016opinion}, natural language inference \citep{bowman2015large}, and semantic textual similarity \citep{agirre2012semeval}. Recently, \citet{devlin2018bert} have shown that fine-tuning a pre-trained language model can achieve state-of-the-art performances
on various text classification tasks without any task-specific architectural adaptations. Thereafter, numerous pre-training and fine-tuning strategies to improve the classification accuracy further have been proposed \citep{liu2019roberta,lan2020albert,sanh2019distilbert,clark2020electra,sun2019fine,mosbach2020stability,zhang2020revisiting}. However, a vast majority of the works have focused on evaluating the accuracy of the models only and overlooked their \textit{reliability} \citep{hendrycks2020pretrained}, \eg, robustness to out-of-distribution (OOD) samples drawn far from the training data (or in-distribution samples).

\begin{figure}[t]
\centering
\includegraphics[width=0.45\textwidth]{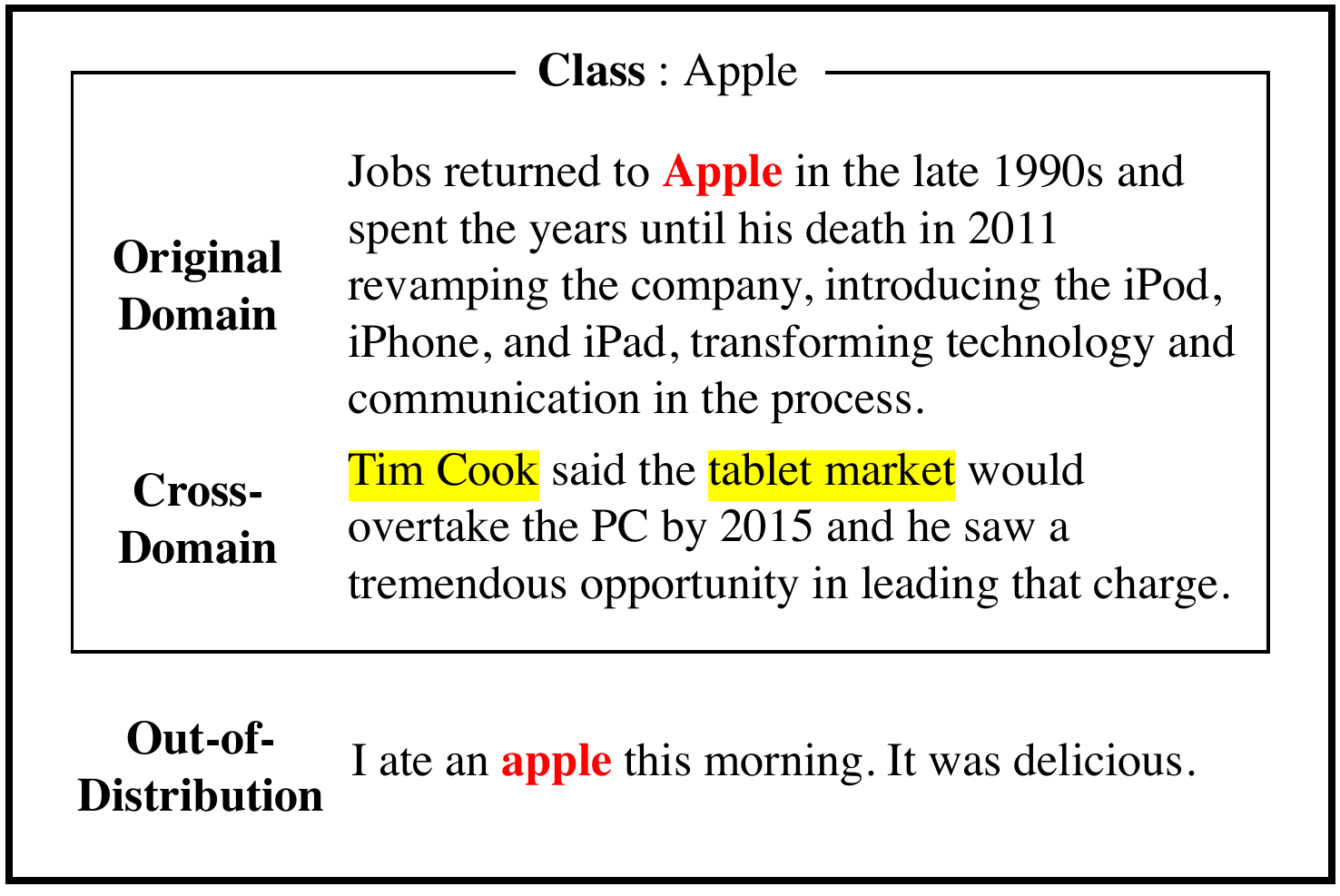}
\caption{
Out-of-distribution (OOD) and cross-domain examples, where class `Apple' is the original domain. The OOD sample contains the word `apple' (red) but in a different context. The cross-domain sample does not share the words (\eg, `Tim Cook') with the original domain, but it still contains some clues (yellow) to guess the correct class.
} \label{fig:intro}
\end{figure}

\begin{figure*}[t]
\centering
\begin{subfigure}{0.33\textwidth}
\includegraphics[width=\textwidth]{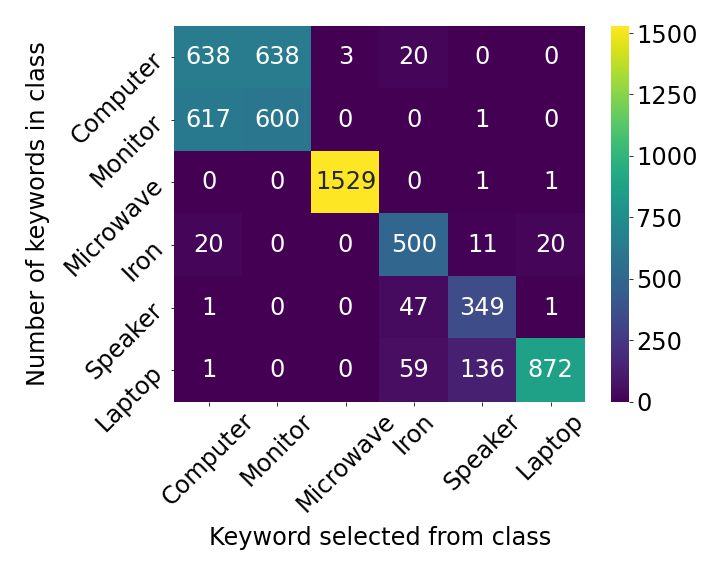}
\caption{In-distribution}
\label{fig:freq_id}
\end{subfigure}
\begin{subfigure}{0.33\textwidth}
\includegraphics[width=\textwidth]{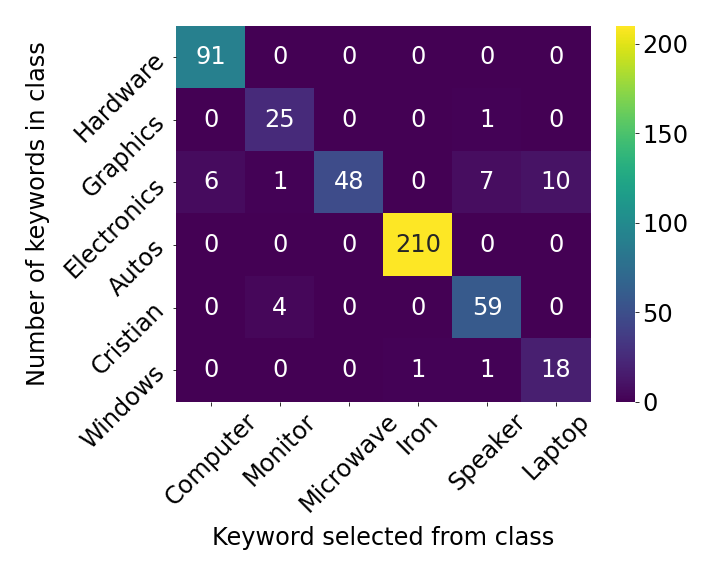}
\caption{Out-of-distribution}
\label{fig:freq_ood}
\end{subfigure}
\begin{subfigure}{0.33\textwidth}
\includegraphics[width=\textwidth]{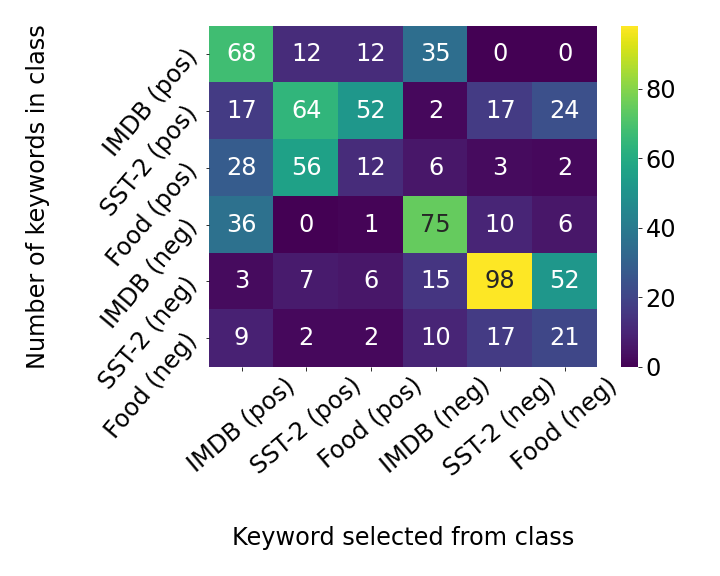}
\caption{Cross-domain}
\label{fig:freq_gen}
\end{subfigure}
\caption{
Frequency of the keywords selected from the source class (x-axis) in the target class (y-axis). (a) Both the source and target classes are in-distribution, (b) source and target distributions are in- and out-of-distribution, respectively, and (c) source and target distributions are identical but of multiple domains. In (b), one can see that OOD classes often contain the same keywords from the (similar but different) in-distribution classes, \eg, see `Iron' and `Autos.' In (c), one can see that the keywords in one domain do not perfectly align to the other domain, \eg, see `IMDB (neg)' and `Food (pos).'
}\label{fig:freq}
\end{figure*}

While \textit{pre-trained} language models are known to be robust in some sense \cite{hendrycks2020pretrained}, we find that \textit{fine-tuned} models suffer from the over-reliance problem, \ie, making predictions based on only a limited number of domain-specific keywords instead of looking at the whole context. For example, consider a classification task of `Apple' visualized in Figure~\ref{fig:intro}. If the most in-distribution samples contain the keyword `Apple,' the fine-tuned model can predict the class solely based on the existence of the keyword. However, a reliable classifier should detect that the sentence ``I ate an apple this morning'' is an out-of-distribution sample \citep{hendrycks2017baseline,shu2017doc,tan2019out}. On the other hand, the sentence ``Tim Cook said that \dots'' should be classified as the topic `Apple' although it does not contain the keyword `Apple' and the keyword `Tim Cook' is not contained in the training samples. In other words, the reliable classifier should learn decision rules that generalize across domains \citep{fei2015social,bhatt2015iterative,bhatt2016cross}.

This problematic phenomenon frequently happens in real-world datasets. To verify this, we extract the keywords from Amazon 50 class reviews \citep{chen2014mining} dataset and sentiment analysis datasets (IMDB \citep{maas2011learning}; SST-2 \citep{socher2013recursive}; Fine Food \citep{mcauley2013amateurs}), following the attention-based scheme illustrated in Section \ref{sec:keyword-selection}. Figure~\ref{fig:freq} shows the frequency of the keywords selected from the source class in the target class. Figure~\ref{fig:freq_id} shows that the keywords are often strongly tied with the class, which leads the model to learn a shortcut instead of the context. Figure~\ref{fig:freq_ood} shows the results where the source and target classes are different classes of the Amazon reviews dataset. Here, OOD classes often contain the same keywords from the in-distribution classes, \eg, the class `Autos' contains the same keywords as the class `Iron.' On the other hand, Figure~\ref{fig:freq_gen} shows the results where both source and target classes are sentiments (`pos' and `neg') classes in IMDB, SST-2, and Fine Food datasets. While the same sentiment shares the same keywords, the alignment is not perfect; \eg, `IMDB (neg)' and `Food (pos)' contain the same keywords.

\subsection{Contribution}

We propose a simple yet effective fine-tuning method coined masked keyword regularization (MASKER), which handles the over-reliance (on keywords) problem and facilitates the context-based prediction. In particular, we introduce two regularization techniques: (a) masked keyword reconstruction and (b) masked entropy regularization. First, (a) forces the model to predict the masked keywords from understanding the context around them. This is inspired by masked language modeling from BERT \citep{devlin2018bert}, which is known to be helpful for learning context. Second, (b) penalizes making high-confidence predictions from ``cut-out-context'' sentences, that non-keywords are randomly dropped, in a similar manner of Cutout \cite{devries2017improved} used for regularizing image classification models. We also suggest two keyword selection schemes, each relying on dataset statistics and attention scores. We remark that all proposed techniques of MASKER can be done in an \textit{unsupervised} manner.

We demonstrate that MASKER, applied to the pre-trained language models: BERT \citep{devlin2018bert}, RoBERTa \citep{liu2019roberta}, and ALBERT \citep{lan2020albert}, significantly improves the OOD detection and cross-domain generalization performance, without degrading the classification accuracy. We conduct OOD detection experiments on 20 Newsgroups \citep{lang1995newsweeder}, Amazon 50 class reviews \citep{chen2014mining}, Reuters \citep{lewis2004rcv1}, IMDB \citep{maas2011learning}, SST-2 \citep{socher2013recursive}, and Fine Food \citep{mcauley2013amateurs} datasets, and cross-domain generalization experiments on sentiment analysis \citep{maas2011learning,socher2013recursive,mcauley2013amateurs}, natural language inference \citep{williams2017broad}, and semantic textual similarity \citep{wang2019glue} tasks. In particular, our method improves the area under receiver operating characteristic (AUROC) of BERT from 87.0\% to 98.6\% for OOD detection under 20 Newsgroups to SST-2 task, and reduce the generalization gap from 19.2\% to 10.9\% for cross-domain generalization under Fine Food to IMDB task.

\subsection{Related Work}

\textbf{Distribution shift in NLP.}
The reliable text classifier should detect distribution shift, \ie, test distribution is different from the training distribution. However, the most common scenarios: OOD detection and cross-domain generalization are relatively under-explored in NLP domains \citep{hendrycks2020pretrained, MarasovicGradient2018NLP}. \citet{hendrycks2020pretrained} found that pre-trained models are robust to the distribution shift compared to traditional NLP models. We find that the pre-trained models are not robust enough, and we empirically show that pre-trained models are still relying on undesirable dataset bias. Our method further improves the generalization performance, applied to the pre-trained models.

\textbf{Shortcut bias.}
One may interpret the over-reliance problem as a type of shortcut bias \citep{geirhos2020shortcut}, \ie, the model learns an easy-to-learn but not generalizable solution, as the keywords can be considered as a shortcut. The shortcut bias is investigated under various NLP tasks \citep{sun2019fine}, \eg, natural language inference \citep{mccoy2019right}, reasoning comprehension \citep{niven2019probing}, and question answering \citep{min2019compositional}. To our best knowledge, we are the first to point out that the over-reliance on keywords can also be a shortcut, especially for text classification. We remark that the shortcut bias is not always harmful as it can be a useful feature for in-distribution accuracy. However, we claim that they can be problematic for unexpected (\ie, OOD) samples, as demonstrated in our experiments.

\textbf{Debiasing methods.}
Numerous debiasing techniques have been proposed to regularize shortcuts, \eg, careful data collection \citep{choi2018quac,reddy2019coqa}, bias-tailored architecture \citep{agrawal2018don}, and adversarial regularization \citep{clark2019don,minderer2020automatic,nam2020learning}. However, most prior work requires supervision of biases, \ie, the shortcuts are explicitly given. In contrast, our method can be viewed as an unsupervised debiasing method, as our keyword selection schemes automatically select the keywords.

\section{Masked Keyword Regularization}
\label{sec:method}

We first introduce our notation and architecture setup; then propose the keyword selection and regularization approaches in Section \ref{sec:keyword-selection} and Section \ref{sec:keyword-masking}, respectively.

\textbf{Notation.}
The text classifier $f: \mathbf{x} \mapsto y$ maps a document $\mathbf{x}$ to the corresponding class $y \in \{1, \dots, C\}$. The document $\mathbf{x}$ is a sequence of tokens $t_{i} \in \mathcal{V}$, \ie, $\mathbf{x} = [t_1, \dots, t_T]$ where $\mathcal{V}$ is the vocabulary set and $T$ is the length of the document. Here, the full corpus $\mathcal{D} = \{(\mathbf{x}, y)\}$ is a collection of all documents, and the class-wise corpus $\mathcal{D}_c = \{(\mathbf{x},y) \in \mathcal{D} \mid y = c\}$ is a subset of $\mathcal{D}$ of class $c$. The keyword set $\mathcal{K} \subset \mathcal{V}$ is the set of vocabularies which mostly affects to the prediction.\footnote{Chosen by our proposed keyword selection (Section \ref{sec:keyword-selection}).} The keyword $\mathbf{k} = [k_1,\dots,k_L]$ of the document $\mathbf{x}$ is given by $\mathbf{k} = [t_i \in \mathbf{x} \mid t_i \in \mathcal{K}]$, where $L \le T$ is the number of keywords in the document $\mathbf{x}$.

\textbf{Architecture.}
We assume the pre-trained language model follows the bi-directional Transformer \citep{vaswani2017attention} architecture, widely used in recent days \citep{devlin2018bert,liu2019roberta,lan2020albert}. They consist of three components: embedding network, document classifier, and token-wise classifier. Given document $\mathbf{x}$, the embedding network produces (a) a document embedding (for an entire document), and (b) token embeddings, which correspond to each input token. The document and token-wise classifier predict the class of document and tokens, respectively, from the corresponding embeddings. For the sake of simplicity, we omit the shared embedding network and denote the document and token-wise classifier as $f_\mathtt{doc}: \mathbf{x} \mapsto y$ and $f_\mathtt{tok}: \mathbf{x} = [t_1,\dots,t_T] \mapsto \mathbf{s} = [s_1,\dots,s_T]$, respectively, where $s_i \in \mathcal{V}$ is a target token corresponds to $t_i$.

\begin{figure}[t]
\centering
\includegraphics[width=0.45\textwidth]{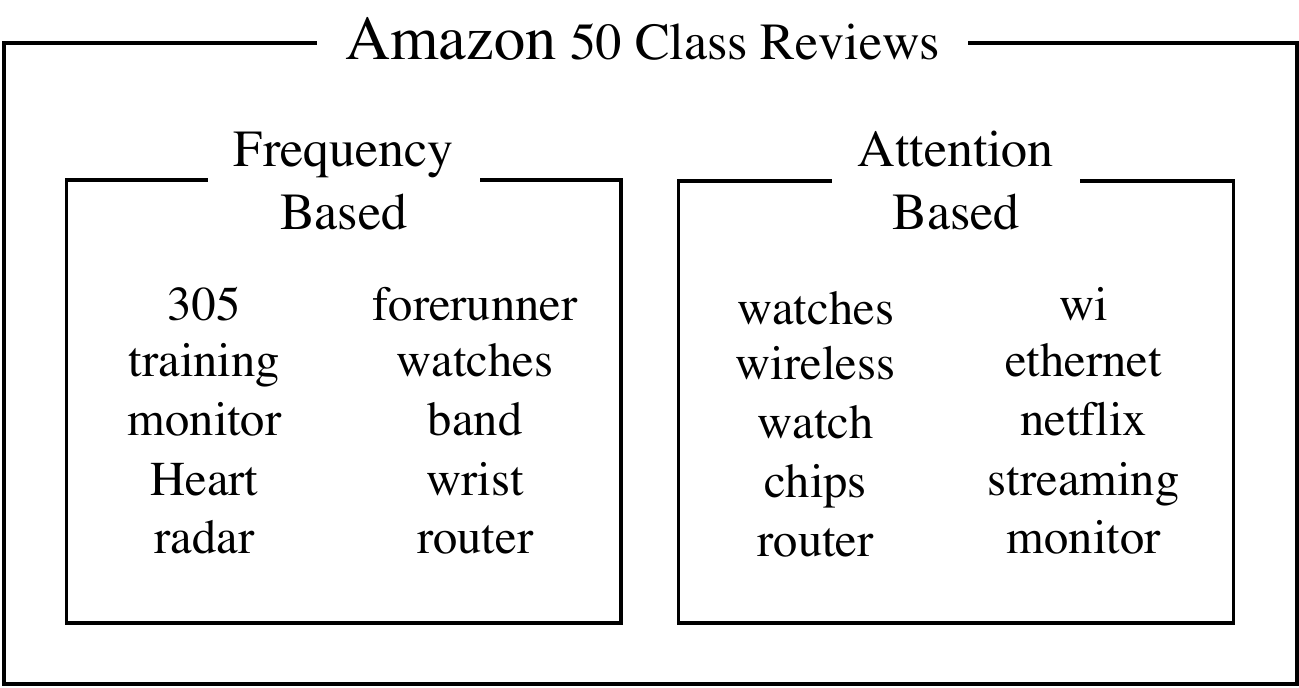}
\caption{
Top 10 keywords chosen from the frequency-based and attention-based selection schemes under the Amazon 50 class reviews dataset. The frequency-based scheme chooses uninformative words (\eg, `305'), while the attention-based scheme chooses more informative ones (\eg, `watch').
}\label{fig:keyword}
\end{figure}

\begin{figure*}[t]
\centering
\begin{subfigure}{0.49\textwidth}
\includegraphics[width=\textwidth]{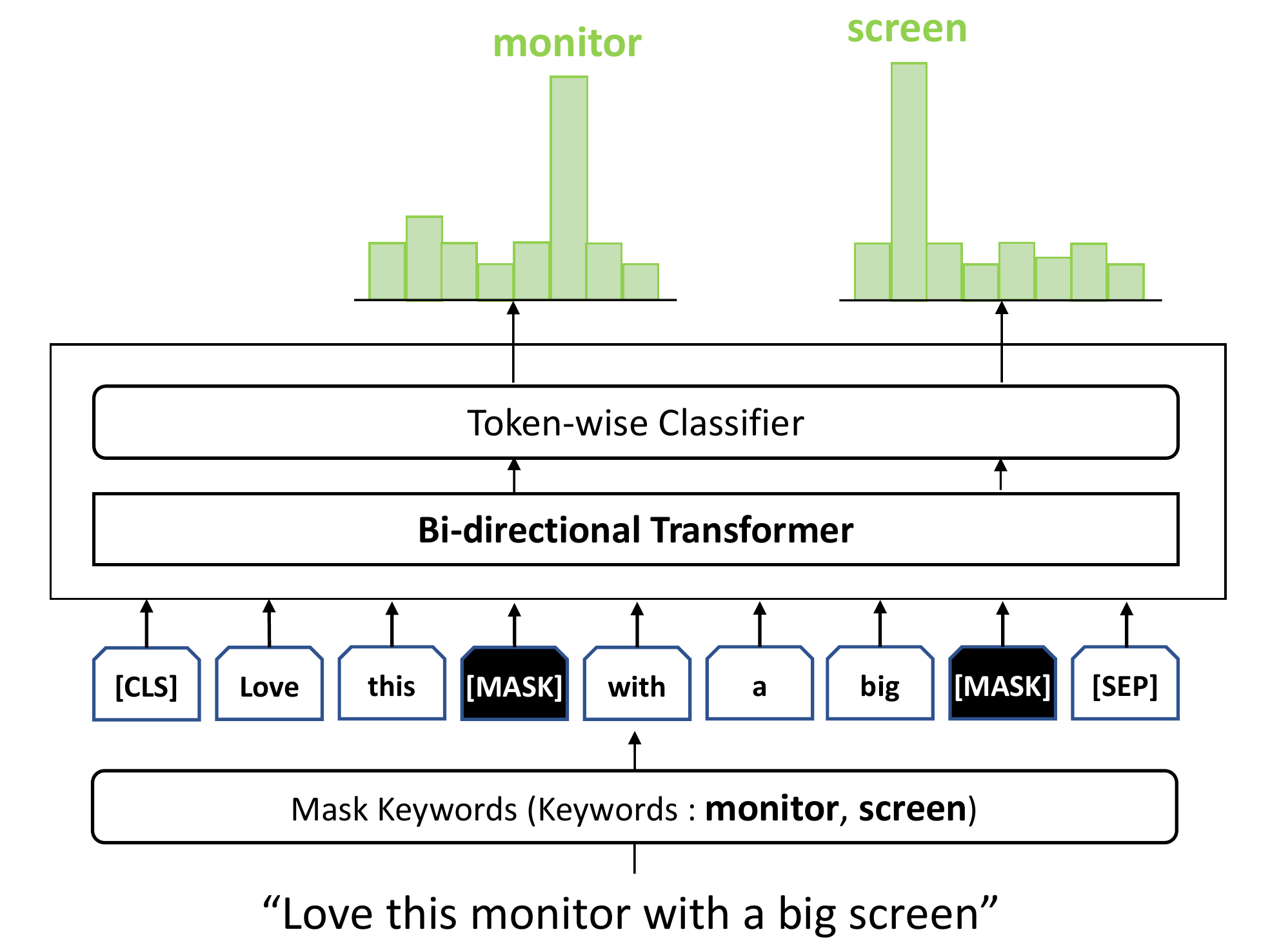}
\caption{Masked keyword reconstruction}
\label{fig:mkr}
\end{subfigure}
\begin{subfigure}{0.49\textwidth}
\includegraphics[width=\textwidth]{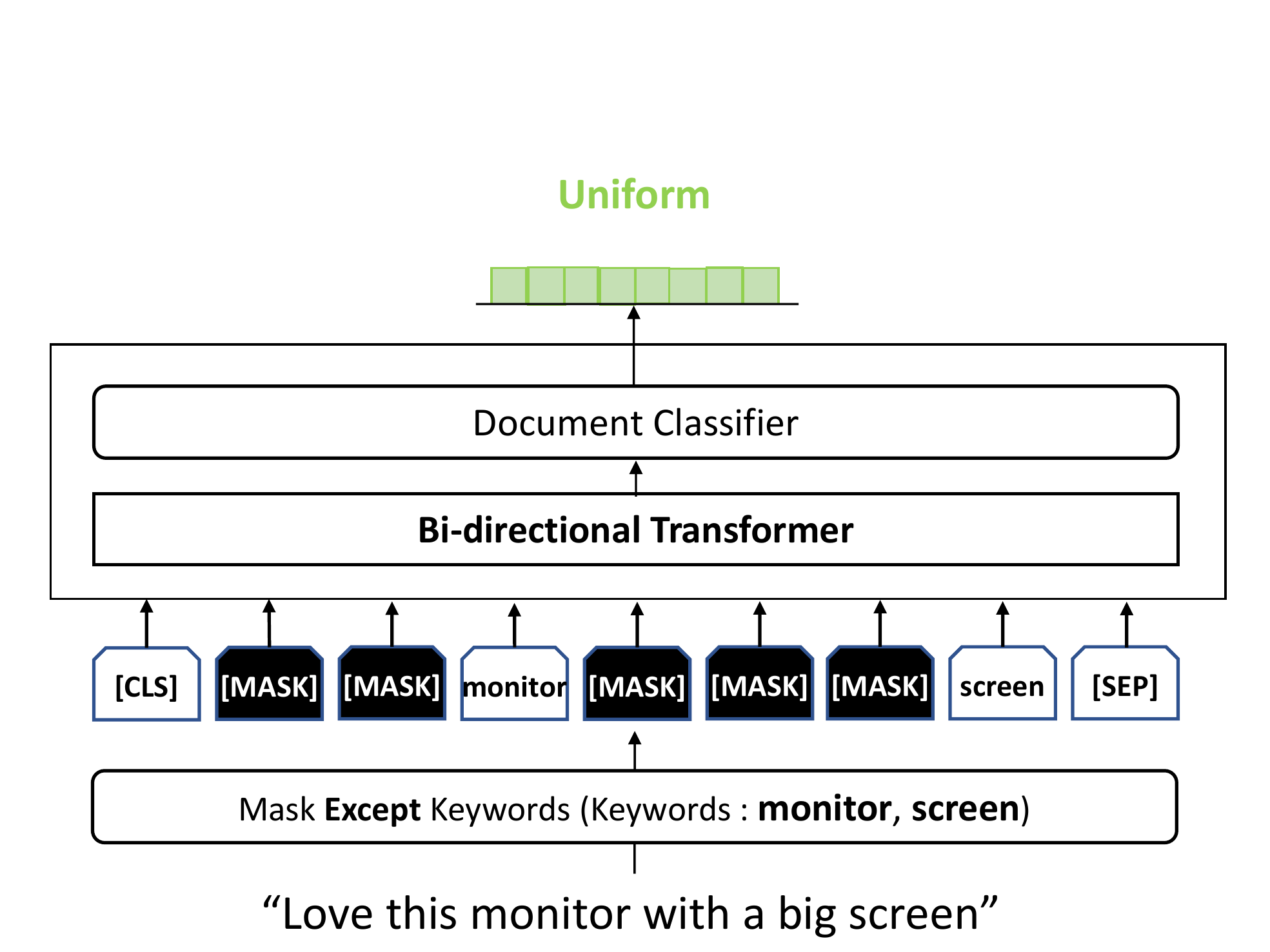}
\caption{Masked entropy regularization}
\label{fig:mer}
\end{subfigure}
\caption{
Illustration of two portions of our proposed method, MASKER: (a) \emph{Masked keyword reconstruction} masks keyword tokens in input sentences and forces the model to predict the original words in masked tokens. (b) \emph{Masked entropy regularization} masks non-keyword tokens in input sentences and forces the model to print uniform distribution, as regarding it as OOD.
}\label{fig:method}
\end{figure*}

\subsection{Keyword Selection Schemes}
\label{sec:keyword-selection}

We consider two keyword selection schemes, based on the dataset statistics (model-free) and trained models. While the former is computationally cheaper, the latter performs better; hence, one can choose its purpose.

\textbf{Frequency-based.}
We first choose the keywords using the relative frequency of the words in the dataset. Specifically, we use the term frequency-inverse document frequency (TF-IDF; \citet{robertson2004understanding}) metric, which measures the importance of the token by comparing the frequency in the target documents (term frequency) and the entire corpus (inverse document frequency). Here, the keywords are defined as the tokens with the highest TF-IDF scores.
Formally, let $\mathbf{X}_c$ be a large document that concatenates all tokens in a class-wise corpus $\mathcal{D}_c$, and $\mathbf{D} = [\mathbf{X}_1,\dots,\mathbf{X}_C]$ be a corpus of such large documents. Then, the frequency-based score of token $t$ is given by
\begin{align}
     s^\mathtt{freq}(t) = \max_{c \in \{1,\dots,C\}} ~\tf(t, \mathbf{X}_c) \cdot \idf(t, \mathbf{D})
    \label{eq:frequency-score}
\end{align}
where $\tf(t, \mathbf{X}) = 0.5 + 0.5\cdot n_{t, \mathbf{X}} / \max\{{n_{t', \mathbf{X}}}: t' \in \mathbf{X}\}$,
$\idf(t,\mathbf{D}) = \log( |\mathbf{D}| / |\{ \mathbf{X} \in \mathbf{D}: t \in \mathbf{X} \}| )$, and $n_{t, \mathbf{x}}$ is number of token $t$ in document $\mathbf{x}$. Note that the frequency-based selection is model-agnostic and easily computed, but does not reflect the contribution of the words to the prediction.

\textbf{Attention-based.}
We also choose the keywords using the model attention as it is a more direct and effective way to measure the importance of words on model prediction. To this end, we first train a model with a standard approach using the cross-entropy loss $\mathcal{L}_\mathtt{CE}$, which leads the model to suffer from the over-reliance (on keywords) issue. Our idea is to use the attention values of the model for choosing the keywords. Here, the keywords are defined as the tokens with the highest attention values. Formally, let $\textbf{a} = [a_1,\dots,a_T] \in \mathbb{R}^T$ be attention values of the document embedding, where $a_i$ corresponds to the input token $t_i$. Then, the attention-based score of token $t$ is given by
\begin{align}
     s^\mathtt{attn}(t) = \sum_{(\mathbf{x},y) \in \mathcal{D}}
        \frac{1}{n_{t, \mathbf{x}}} \sum_{i \in \{1,\dots,T\}} \mathbb{I}(t_i = t) \cdot \frac{a_i}{\norm{\textbf{a}}}
    \label{eq:attention-score}
\end{align}
where $\mathbb{I}$ is an indicator function and $\norm{\cdot}$ is $\ell_2$-norm.

We choose the keywords by picking the top $K$ tokens according to the scores in Eq. \eqref{eq:frequency-score} and Eq. \eqref{eq:attention-score} for each selection scheme, respectively. We also test the class-balanced version, \ie, pick the top $K/C$ tokens for each class, but the class-agnostic one performed better.

\textbf{Comparison of the selection schemes.}
We observe that the frequency-based scheme often selects uninformative keywords that uniquely appears in some class. In contrast, the attention-based scheme selects more general keywords that actually influence the prediction. Figure \ref{fig:keyword} shows the keywords chosen by both selection schemes: the frequency-based scheme chooses uninformative words such as `305' and `forerunner,' while the attention-based scheme chooses more informative ones such as `watch' or `chips.'

\subsection{Regularization via Keyword Masking}
\label{sec:keyword-masking}

Using the chosen keywords, we propose two regularization techniques to reduce the over-reliance issue and facilitate the model to look at the contextual information.

\textbf{Masked keyword reconstruction.}
To enforce the model to look at the surrounding context, we guide the model to reconstruct the keywords from keyword-masked documents. Note that it resembles the masked language model \citep{devlin2018bert}, but we mask the \textit{keywords} instead of random words. Masked keyword reconstruction only regularizes sentences with keywords, and we omit the loss for ones without any keywords. Formally, let $\tilde{\mathbf{k}}$ be a random subset of the full keyword $\mathbf{k}$ (selected as in Section~\ref{sec:keyword-selection}), that each element is chosen with probability $p$ independently. We mask $\tilde{\mathbf{k}}$ from the original document $\mathbf{x}$ and get the masked document $\tilde{\mathbf{x}} = \mathbf{x} - \tilde{\mathbf{k}}$. Then, the masked keyword reconstruction (MKR) loss is
\begin{align}
    \mathcal{L}_\mathtt{MKR}(\tilde{\mathbf{x}},v)
    := \sum_{i \in \mathtt{index}(\tilde{\mathbf{k}})} \mathcal{L}_\mathtt{CE}(f_\mathtt{tok}(\tilde{\mathbf{\mathbf{x}}})_i, v_i)
\end{align}
where $\mathtt{index}(\tilde{\mathbf{k}})$ is the index of the keywords $\tilde{\mathbf{k}}$
with respect to the original document $\mathbf{x}$, $v_i$ is the index of the keywords with respect to the vocabulary set. We remark that the \textit{reconstruction} part is essential; we also test simply augmenting the masked documents, \ie, $\mathcal{L}_\mathtt{CE}(f_\mathtt{doc}(\tilde{\mathbf{x}}),y)$, but it performed worse. Choosing proper keywords is also crucial; attention-based keywords performs better than frequency-based or random keywords, as shown in Table~\ref{tab:ood_ablation} and Table~\ref{tab:gen_ablation}.

\textbf{Masked entropy regularization.}
Furthermore, we regularize the prediction of the \textit{context}-masked documents, that context (non-keyword) words are randomly dropped. The model should not classify the context-masked documents correctly as they lost the original context. Formally, let $\widehat{\mathbf{c}}$ be a randomly chosen subset of the full context words $\mathbf{c} = \mathbf{x} - \mathbf{k}$, where each element is chosen with probability $q$ independently. We mask $\widehat{\mathbf{c}}$ from the original document $\mathbf{x}$ and get the context-masked document $\widehat{\mathbf{x}} = \mathbf{x} - \widehat{\mathbf{c}}$. Then, the masked entropy regularization (MER) loss is
\begin{align}
    \mathcal{L}_\mathtt{MER}(\widehat{\mathbf{x}}) 
    := D_\mathrm{KL}( \mathcal{U}(y) || f_\mathtt{doc}(\widehat{\mathbf{x}}) )
\end{align}
where $D_\mathrm{KL}$ is the KL-divergence and $\mathcal{U}(y)$ is a uniform distribution. We remark that MER does not degrade the classification accuracy since it regularizes non-realistic context-masked sentences, rather than full documents. Table~\ref{tab:ood_ablation} shows that MER does not drop the classification accuracy in original domain, while Table~\ref{tab:gen_ablation} and Table~\ref{tab:gen_main} show that MER improves the cross-domain accuracy. On the other hand, MER differs from the prior sentence-level objectives, \eg, next sentence prediction \citep{devlin2018bert}, as our goal is to regularize shortcuts, not learning better in-domain representation.

To sum up, the final objective is given by
\begin{align}
    \mathcal{L}_\mathtt{total}
    = \mathcal{L}_\mathtt{CE} + \lambda_\mathtt{MKR} \mathcal{L}_\mathtt{MKR} + \lambda_\mathtt{MER} \mathcal{L}_\mathtt{MER}
\end{align}
where $\lambda_\mathtt{MKR}$ and $\lambda_\mathtt{MER}$
are hyperparameters for the MKR and MER losses, respectively. Figure~\ref{fig:method} visualizes the proposed losses, and the overall procedure is in Appendix~\ref{app:overall}.

\begin{table*}[t]
\centering\small
\begin{tabular}{cccccccc|c}
\toprule
Method & Classifier & \makecell{Keyword} & MKR & MER & AUROC $\uparrow$ & EER $\downarrow$ & \makecell{Detection\\Accuracy $\uparrow$} & \makecell{Classification\\Accuracy $\uparrow$} \\
\midrule
OC-SVM  & -       & -    & -   & -    & 57.01\stdv{1.08} & 41.12\stdv{0.90} & 72.99\stdv{4.69} & - \\
OpenMax & -       & -    & - & - & 53.02\stdv{3.74} & 55.21\stdv{2.03} & 74.01\stdv{0.58} & 79.01\stdv{0.51} \\
DOC     & -       & -    & - & - & 75.12\stdv{3.06} & 25.55\stdv{4.12} & 80.21\stdv{0.76} & 83.14\stdv{0.82} \\ 
\midrule
\multirow{2}{*}{BERT}
    & Multi-class & -    & - & - & 78.46\stdv{1.16} & 28.30\stdv{1.11} & 81.19\stdv{0.62} & 84.76\stdv{0.35} \\
    & 1-vs-rest      & -    & - & - & 80.56\stdv{0.84} & 25.71\stdv{0.71} & 80.78\stdv{0.48} & 84.66\stdv{0.26} \\
\midrule
\multirow{7}{*}{\makecell{BERT\\+MASKER\\(ours)}}
    & 1-vs-rest & Random & \checkmark & - & 80.52\stdv{0.72} & 25.42\stdv{0.38} & 80.83\stdv{0.33} & 85.06\stdv{0.14} \\
    & 1-vs-rest & Random & - & \checkmark & 81.51\stdv{0.27} & 24.17\stdv{0.87} & 82.13\stdv{0.58} & 84.89\stdv{0.71} \\
    \cmidrule(lr){2-9}
    & 1-vs-rest & Frequency & \checkmark & - & 81.32\stdv{0.78} & 25.12\stdv{1.18} & 81.33\stdv{0.67} & 83.35\stdv{1.77} \\
    & 1-vs-rest & Frequency & - & \checkmark & 82.54\stdv{0.54} & 23.88\stdv{0.76} & 82.72\stdv{0.58} & 85.25\stdv{0.70} \\
    \cmidrule(lr){2-9}
    & 1-vs-rest & Attention & \checkmark & - & 83.33\stdv{0.44} & 24.55\stdv{0.67} & 82.11\stdv{0.68} & \textbf{85.27\stdv{0.33}} \\
    & 1-vs-rest & Attention & - & \checkmark & 82.60\stdv{0.51} & 25.02\stdv{0.83} & 80.99\stdv{0.65} & 85.02\stdv{0.29} \\
    & 1-vs-rest & Attention & \checkmark & \checkmark & \textbf{85.48\stdv{0.54}} & \textbf{22.30\stdv{1.20}} & \textbf{85.02\stdv{0.55}} & 85.15\stdv{0.95} \\
\bottomrule
\end{tabular}
\caption{
Ablation study on OOD detection under the Amazon 50 class reviews. We use 25\% of classes as in-distribution, and the rest as OOD. The reported results are averaged over five trials, subscripts denote standard deviations, and the best results are highlighted in bold. All components of our method contribute to the OOD detection performance (\%).
}\label{tab:ood_ablation}
\end{table*}
\begin{table*}[t]
\centering\small
\begin{tabular}{cccc|c|c|c}
\toprule
\multirow{2}{*}{ID} & \multirow{2}{*}{OOD} & \multirow{2}{*}{OC-SVM} & \multirow{2}{*}{DOC} & \multicolumn{3}{c}{Vanilla / Residual / MASKER} \\
& & & & BERT & RoBERTa & ALBERT \\
\cmidrule(lr){1-7}
\multirow{5}{*}{Newsgroup} & Amazon & 62.1 & 84.1 & 85.4/86.7/\textbf{87.0}(+1.9\%) & 85.3/85.9/\textbf{87.2}(+2.3\%) & 86.7/85.4/\textbf{89.4}(+3.1\%) \\
& Reuters & 53.9 & 60.0 & 91.8/93.0/\textbf{97.7}(+6.4\%) & 93.1/92.1/\textbf{93.9}(+0.8\%) & 93.3/92.0/\textbf{94.7}(+1.6\%) \\
& IMDB & 59.8 & 88.6 & 94.6/95.7/\textbf{98.5}(+4.1\%) & 95.2/95.9/\textbf{97.7}(+2.7\%) & 94.5/92.6/\textbf{96.6}(+2.2\%) \\
& SST-2 & 63.0 & 88.1 & 87.0/97.0/\textbf{98.6}(+13.3\%) & 94.7/94.9/\textbf{98.2}(+3.6\%) & 95.8/95.6/\textbf{98.1}(+2.4\%) \\
& Fine Food & 62.8 & 81.3 & 85.3/87.3/\textbf{93.4}(+9.5\%) & 88.7/89.4/\textbf{92.9}(+4.7\%) & 77.6/86.6/\textbf{91.6}(+18.1\%) \\
\cmidrule(lr){1-7}
\multirow{5}{*}{Amazon} & Newsgroup & 61.3 & 81.3 & 84.8/83.9/\textbf{87.2}(+2.8\%) & 87.9/86.6/\textbf{91.0}(+3.5\%) & 87.3/85.0/\textbf{88.4}(+1.3\%) \\
& Reuters & 55.5 & 79.8 & 89.7/89.7/\textbf{93.5}(+4.2\%) & 92.3/92.6/\textbf{93.6}(+1.4\%) & 93.1/93.5/\textbf{94.5}(+1.5\%) \\
& IMDB & 66.2 & 89.6 & 93.3/92.8/\textbf{95.2}(+2.0\%) & 90.1/87.0/\textbf{93.3}(+3.5\%) & 89.9/88.6/\textbf{95.6}(+6.4\%) \\
& SST-2 & 60.9 & 91.5 & 93.0/88.9/\textbf{95.6}(+2.8\%) & 92.4/94.8/\textbf{96.4}(+4.4\%) & 93.4/91.9/\textbf{96.9}(+3.7\%) \\
& Fine Food & 51.1 & 66.8 & 78.5/77.7/\textbf{84.9}(+8.2\%) & 74.9/80.0/\textbf{80.7}(+7.8\%) & 82.6/86.3/\textbf{87.3}(+5.7\%) \\
\bottomrule
\end{tabular}
\caption{
AUROC (\%) on various OOD detection scenarios. The reported results are averaged over three trials, and the best results are highlighted in bold. Bracket denotes the relative gain of MASKER over the vanilla model.
}\label{tab:ood_main}
\end{table*}
\begin{table*}[ht]
\centering\small
\resizebox{\textwidth}{!}{
\begin{tabular}{ccccccccccc}
\toprule
\multirow{3}{*}{Method} & \multirow{3}{*}{Classifier} & \multirow{3}{*}{Keyword} & \multirow{3}{*}{MKR} & \multirow{3}{*}{MER} & \multicolumn{6}{c}{Dataset (Train $\to$ Test)} \\
\cmidrule(lr){6-11}
& & & & & \makecell{IMDB\\$\to$ SST-2} & \makecell{IMDB\\$\to$ Food} & \makecell{SST-2\\$\to$ IMDB} & \makecell{SST-2\\$\to$ Food} & \makecell{Food\\$\to$ SST-2} & \makecell{Food\\$\to$ IMDB} \\
\midrule
OpenMax & - & - & - & - & \cell{79.55}{0.78}{-8.12} & \cell{75.41}{1.20}{-12.25} &
\cell{75.30}{0.44}{-7.61} & \cell{62.19}{3.06}{-20.72} & \cell{61.85}{0.63}{-31.70} & \cell{67.50}{1.50}{-26.04}\\
DOC     & - & - & - & - & \cell{77.90}{1.22}{-10.06} & \cell{78.33}{1.52}{-9.64} & \cell{76.88}{0.70}{-6.23} & \cell{64.47}{2.52}{-18.63} & \cell{62.00}{0.86}{-31.27} & \cell{67.31}{1.28}{-25.96} \\
\midrule
\multirow{4}{*}{BERT}
    & Multi-class & - & - & - & \cell{85.92}{1.92}{-7.57} & \cell{92.90}{2.47}{-0.60} & \cell{85.74}{0.56}{-6.74} & \cell{87.57}{1.13}{-4.91} & \cell{67.55}{5.27}{-28.92} & \cell{77.31}{2.09}{-19.16} \\
    & 1-vs-rest   & - & - & - & \cell{84.28}{0.23}{-8.92} & \cell{87.81}{3.91}{-5.39} & \cell{85.34}{0.63}{-7.46} & \cell{84.35}{1.48}{-8.45} & \cell{64.57}{1.27}{-32.15} & \cell{81.34}{0.78}{-12.16}\\
\midrule
\multirow{14}{*}{\makecell{BERT\\+MASKER\\(ours)}}
    & 1-vs-rest & Random & \checkmark & - & \cell{87.29}{1.48}{-6.29} & \cell{90.52}{1.28}{-3.06} & \cell{86.57}{0.87}{-7.60} & \cell{78.00}{0.86}{-16.17} & \cell{78.79}{1.15}{-17.51} & \cell{84.56}{1.59}{-12.91}\\
    & 1-vs-rest & Random & - & \checkmark & \cell{86.84}{1.76}{-5.97} & \cell{90.27}{1.25}{-2.54} & \cell{87.18}{0.81}{-8.52} & \cell{85.91}{1.04}{-9.79} & \cell{79.50}{0.46}{-17.03} & \cell{84.61}{0.41}{-11.92}\\
    \cmidrule(lr){2-11}
    & 1-vs-rest & Frequency & \checkmark & - & \cell{86.52}{1.40}{-6.04} & \cell{88.41}{1.72}{-4.15} & \cell{87.06}{1.14}{-7.37} & \cell{79.99}{1.88}{-14.44} & \cell{74.72}{1.90}{-23.73} & \cell{80.94}{2.80}{-17.51}\\
    & 1-vs-rest & Frequency & - & \checkmark & \cell{86.38}{0.88}{-6.72} & \cell{84.20}{2.59}{-8.90} & \cell{85.31}{2.31}{-13.55} & \cell{88.43}{1.85}{-10.43} & \cell{75.34}{1.54}{-20.66} & \cell{85.34}{0.99}{-10.66} \\
    \cmidrule(lr){2-11}
    & 1-vs-rest & Attention & \checkmark & - & \cell{87.50}{1.49}{-5.91} & \cell{92.03}{2.97}{-5.74} & \cell{87.78}{1.34}{-4.92} & \textbf{\cell{90.12}{2.68}{-2.58}} & \cell{75.57}{4.02}{-20.82} & \cell{79.32}{5.09}{-17.07} \\
    & 1-vs-rest & Attention & - & \checkmark & \cell{87.71}{0.71}{-5.59} & \cell{90.39}{0.39}{-2.63} & \cell{84.92}{2.52}{-7.64} & \cell{87.21}{1.01}{-5.35} & \cell{75.80}{1.84}{-20.87} & \cell{82.13}{2.39}{-14.54}\\
    & 1-vs-rest & Attention & \checkmark & \checkmark & \textbf{\cell{88.02}{1.31}{-5.44}} & \textbf{\cell{93.58}{2.63}{+0.12}} & \textbf{\cell{88.43}{0.38}{-3.89}} & \cell{89.21}{0.40}{-3.11} & \textbf{\cell{80.02}{1.52}{-16.44}} & \textbf{\cell{85.57}{0.28}{-10.90}} \\
\bottomrule
\end{tabular}}
\caption{
Ablation study on cross-domain generalization under sentiment analysis task. The reported results are averaged over five trials, subscripts denote standard deviations, bracketed numbers denote the generalization gap from the training domain accuracy, and the best accuracies are highlighted in bold. All components of our method contribute to the cross-domain accuracy (\%).
}\label{tab:gen_ablation}
\end{table*}
\begin{table*}[ht]
\centering\small
\begin{subtable}{\textwidth}
\centering\small
\begin{tabular}{cc|cc|c|c|c}
\toprule
\multirow{2}{*}{Train} & \multirow{2}{*}{Test} & \multirow{2}{*}{OpenMax} & \multirow{2}{*}{DOC} & \multicolumn{3}{c}{Vanilla / Residual / MASKER}\\
& & & & BERT & RoBERTa & ALBERT \\ \cmidrule{1-7}
\multirow{3}{*}{IMDB} & IMDB & 87.7 & 88.0 & 93.5/92.8/93.5\phantom{(+x.x\%)} & 95.3/94.0/95.6\phantom{(+x.x\%)} & 91.6/91.4/90.8\phantom{(+x.x\%)} \\
& SST-2 & 79.6 & 77.9 & 85.9/86.9/\textbf{88.1}(+2.6\%) & 89.7/90.2/\textbf{91.8}(+2.3\%) & 89.8/89.0/\textbf{89.9}(+0.1\%) \\
& Fine Food & 75.4 & 78.3 & 92.9/92.5/\textbf{93.6}(+0.8\%) & 92.6/87.7/\textbf{93.0}(+0.4\%) & 87.1/87.8/\textbf{92.1}(+5.7\%)\\
\cmidrule{1-7}
\multirow{3}{*}{SST-2} & SST-2 & 82.9 & 83.1 & 92.5/90.3/92.3\phantom{(+x.x\%)} & 94.5/92.0/94.3\phantom{(+x.x\%)} & 91.9/90.9/91.5\phantom{(+x.x\%)} \\
& IMDB & 75.3 & 76.9 & 85.7/86.2/\textbf{88.4}(+3.2\%) & 86.0/85.7/\textbf{87.3}(+1.5\%) & 83.0/83.0/\textbf{83.8}(+1.0\%) \\
& Fine Food & 62.2 & 64.5 & 87.6/88.2/\textbf{89.2}(+1.8\%) & 86.5/87.8/\textbf{89.6}(+3.6\%) & 79.0/80.6/\textbf{84.9}(+7.5\%) \\ \cmidrule{1-7}
\multirow{3}{*}{Fine Food} & Fine Food & 93.6 & 93.3 & 96.5/94.8/96.5\phantom{(+xx.x\%)} & 96.9/95.7/97.1\phantom{(+x.x\%)} & 95.5/95.4/96.6\phantom{(+xx.x\%)} \\
& IMDB & 67.5 & 67.3 & 77.3/81.1/\textbf{85.6}(+10.7\%) & 84.1/84.6/\textbf{86.6}(+2.9\%) & 74.7/80.4/\textbf{84.8}(+13.5\%) \\
& SST-2 & 61.9 & 62.0 & 67.6/67.8/\textbf{80.0}(+18.3\%) & 78.5/80.1/\textbf{83.8}(+6.8\%) & 71.7/73.2/\textbf{83.3}(+16.2\%) \\
\bottomrule
\end{tabular}
\caption{Sentiment analysis}\label{tab:gen_sent}
\vspace{0.05in}
\end{subtable}
\begin{subtable}{0.41\textwidth}
\centering\small
\begin{tabular}{c|ccc|}
\toprule
\multirow{3}{*}{Model} & \multicolumn{3}{c|}{Telephone} \\
 & \makecell{Telephone\\(ID)} & \makecell{Letters\\(OOD)} & \makecell{Face-to-Face\\(OOD)} \\ \midrule
    BERT & 80.5 & 77.4 & 77.5 \\
    +MASKER & 80.2 & \makecell{\textbf{80.4}\\(+3.9\%)} & \makecell{\textbf{78.5}\\(+1.3\%)} \\ \midrule
    RoBERTa & 84.8 & 83.6 & 83.5 \\
    +MASKER & 86.0 & \makecell{\textbf{85.9}\\(+2.8\%)} & \makecell{\textbf{83.8}\\(+0.4\%)} \\ \midrule
    ALBERT & 82.9 & 82.8 & 80.9 \\
    +MASKER & 82.5 & \makecell{\textbf{84.0}\\(+1.5\%)} & \makecell{\textbf{86.8}\\(+7.3\%)} \\
\bottomrule
\end{tabular}
\caption{Natural language inference}\label{tab:gen_nli}
\end{subtable}
\begin{subtable}{0.52\textwidth}
\centering\small
\begin{tabular}{ccc|ccc}
\toprule
\multicolumn{3}{c}{MSRvid} & \multicolumn{3}{c}{Images} \\
\makecell{MSRvid\\(ID)} & \makecell{Images\\(OOD)} & \makecell{Headlines\\(OOD)} & \makecell{Images\\(ID)} & \makecell{MSRvid\\(OOD)} & \makecell{Headlines\\(OOD)} \\ \midrule
91.5 & 82.0 & 61.7 & 88.0 & 89.7 & 73.9 \\
91.2 & \makecell{\textbf{84.3}\\(+2.8\%)} & \makecell{\textbf{66.7}\\(+8.1\%)} & 88.1 & \makecell{\textbf{91.6}\\(+2.1\%)} & \makecell{\textbf{75.3}\\(+1.9\%)}\\ \midrule
94.2 & \textbf{88.0} & 80.3 & 91.8 & 92.9 & 84.1 \\
93.7 & \makecell{\textbf{88.0}\\(+0.0\%)} & \makecell{\textbf{84.0}\\(+4.6\%)} & 91.3 & \makecell{\textbf{94.1}\\(+1.3\%)} & \makecell{\textbf{85.3}\\(+1.4\%)}\\ \midrule
92.6 & 81.2 & 60.6 & 90.4 & 90.9 & 69.8 \\
93.3 & \makecell{\textbf{82.6}\\(+1.7\%)} & \makecell{\textbf{68.8}\\(+13.5\%)} & 90.5 & \makecell{\textbf{92.0}\\(+1.2\%)} & \makecell{\textbf{78.4}\\(+12.3\%)}\\
\bottomrule
\end{tabular}
\caption{Semantic textual similarity}\label{tab:gen_stsb}
\end{subtable}
\caption{
Accuracy (\%) of original domain and cross-domain on (a) sentiment analysis, (b) natural language inference, and (c) semantic textual similarity tasks, respectively. The reported results are averaged over three trials for sentiment analysis and semantic textual similarity, and a single trial for natural language inference. Bold denotes the best results among the three methods, and bracket denotes the relative gain of MASKER over the vanilla model.
}\label{tab:gen_main}
\end{table*}

\section{Experiments}
\label{sec:experiments}

We demonstrate the effectiveness of our proposed method, MASKER. In Section \ref{sec:exp_setup}, we describe the experimental setup. In Section \ref{sec:exp_ood} and \ref{sec:exp_gen}, we present the results on OOD detection and cross-domain generalization, respectively.

\subsection{Experimental setup}
\label{sec:exp_setup}

We demonstrate the effectiveness of MASKER, applied to the pre-trained models: BERT \citep{devlin2018bert}, RoBERTa \citep{liu2019roberta} and ALBERT \citep{lan2020albert}. We choose $10 \times C$ keywords in a class agnostic way, where $C$ is the number of classes. We drop the keywords and contexts with probability $p=0.5$ and $q=0.9$ for all our experiments. We use $\lambda_\mathtt{MKR}=0.001$ and $\lambda_\mathtt{MER}=0.001$ for OOD detection, and same $\lambda_\mathtt{MKR}=0.001$ but $\lambda_\mathtt{MER}=0.0001$ for cross-domain generalization, as the entropy regularization gives more gain for reliability than accuracy \citep{pereyra2017regularizing}. We modify the hyperparameter settings of the pre-trained models \citep{devlin2018bert,liu2019roberta}, specified in Appendix~\ref{app:detail}.

\textbf{1-vs-rest classifier.}
Complementary to MASKER, we use 1-vs-rest classifier \citep{shu2017doc} as it further improves the reliability (see Table~\ref{tab:ood_ablation} and Table~\ref{tab:gen_ablation}). Intuitively, 1-vs-rest classifier can reject all classes (all prediction scores are low); hence detect OOD samples well.

\textbf{Baselines.}
We mainly compare MASKER with vanilla fine-tuning of pre-trained models \citep{hendrycks2020pretrained}, with extensive ablation study (see Table~\ref{tab:ood_ablation} and Table~\ref{tab:gen_ablation}). Additionally, we compare with residual ensemble \citep{clark2019don}, applied to the same pre-trained models. Residual ensemble trains a debiased model by fitting the residual from a biased model. We construct a biased dataset by subsampling the documents that contain keywords. To benchmark the difficulty of the task, we also report the classic non-Transformer models, \eg, one-class support vector machine (OC-SVM, \citet{scholkopf2000support}), OpenMax \citep{bendale2016towards}, and DOC \citep{shu2017doc}.

\subsection{OOD Detection}
\label{sec:exp_ood}

We use the highest softmax (or sigmoid) output of the model as confidence score for OOD detection task. We use 20 Newsgroups \citep{lang1995newsweeder} and Amazon 50 class reviews \citep{chen2014mining} datasets for in-distribution, and Reuters \citep{lewis2004rcv1}, IMDB \citep{maas2011learning}, and SST-2 \citep{socher2013recursive} datasets for out-of-distribution.

Table~\ref{tab:ood_ablation} shows an ablation study on MASKER under the Amazon reviews dataset with a split ratio of 25\%. All components of MASKER contribute to OOD detection. Note that MASKER does not degrade the classification accuracy while improving OOD detection. Also, the attention-based selection performs better than the frequency-based or random selection, which implies the importance of selecting suitable keywords. Recall that the attention-based scheme selects the keywords that contribute to the prediction, while the frequency-based scheme often chooses domain-specific keywords that are not generalizable across domains.

Table~\ref{tab:ood_main} shows the results on various OOD detection scenarios, comparing the vanilla fine-tuning, residual ensemble, and MASKER. Notably, MASKER shows the best results in all cases. In particular, MASKER improves the area under receiver operating characteristic (AUROC) from 87.0\% to 98.6\% on 20 Newsgroups to SST-2 task. We find that residual ensemble shows inconsistent gains: it often shows outstanding results (\eg, Newsgroup to SST-2) but sometimes fails (\eg, Amazon to Fine Food). In contrast, MASKER shows consistent improvement over the vanilla fine-tuning.

\begin{figure*}[hbt!]
\centering
\begin{subfigure}{0.24\textwidth}
\includegraphics[width=\textwidth]{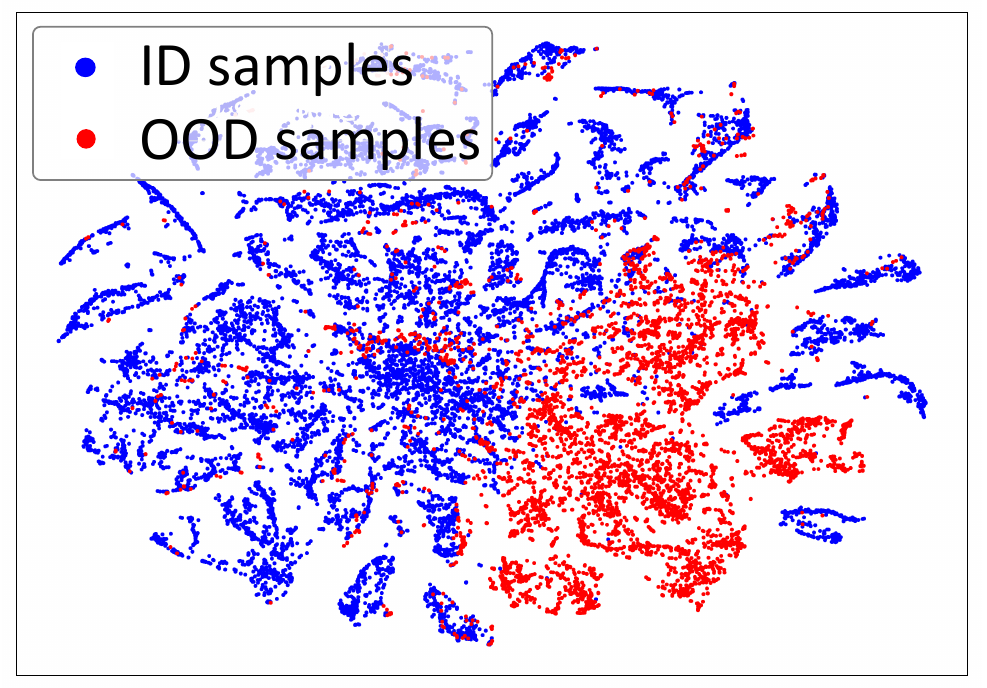}
\caption{BERT}
\label{fig:tsne_a}
\end{subfigure}~
\begin{subfigure}{0.24\textwidth}
\includegraphics[width=\textwidth]{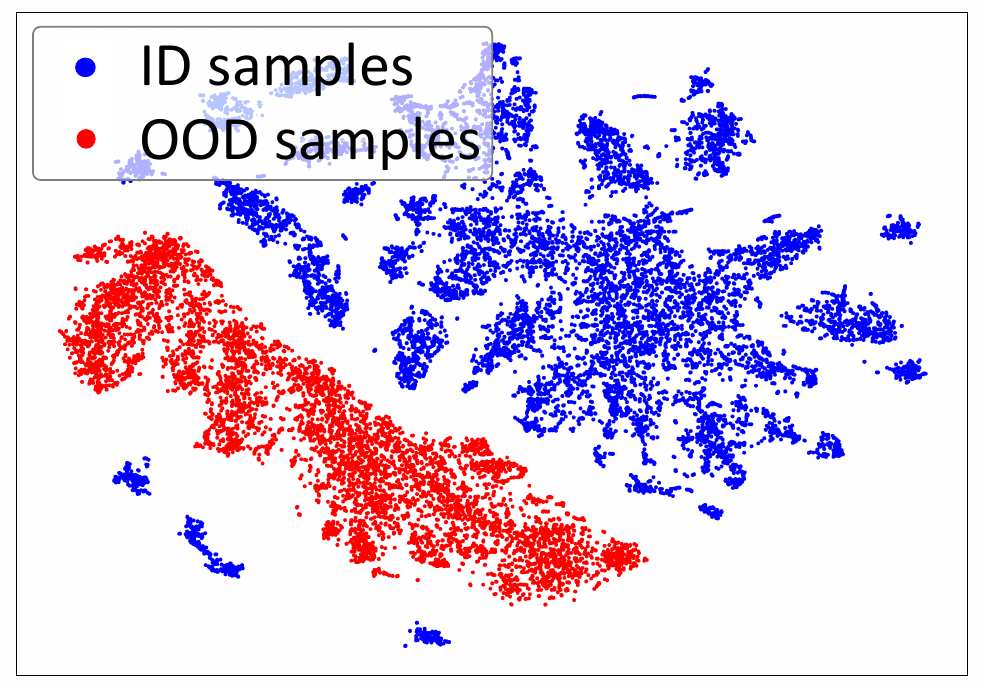}
\caption{MASKER (ours)}
\label{fig:tsne_b}
\end{subfigure}~
\begin{subfigure}{0.24\textwidth}
\includegraphics[width=\textwidth]{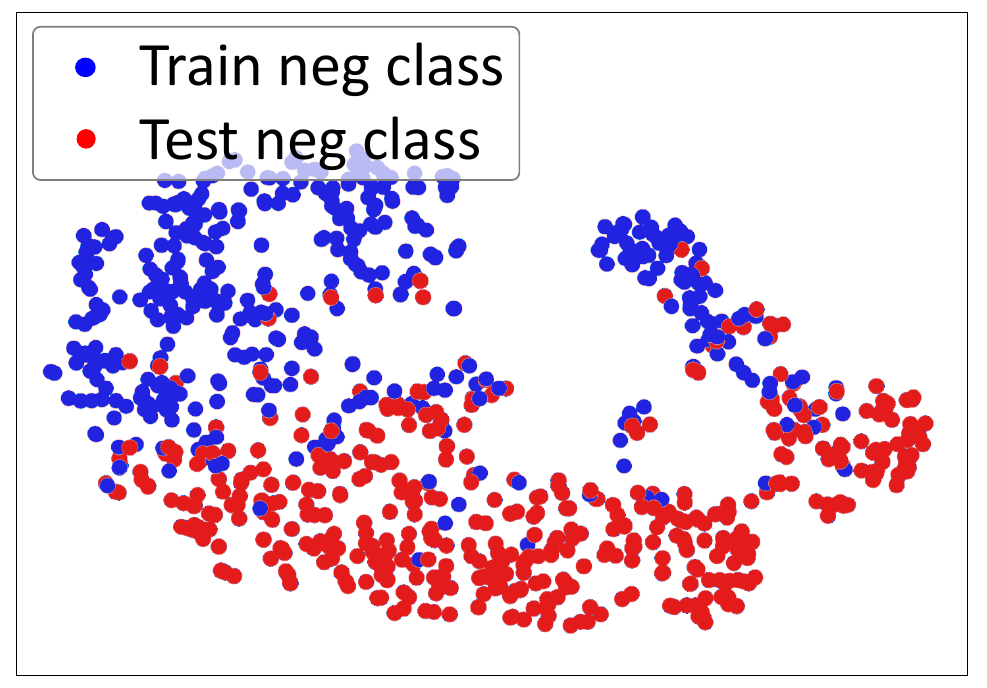}
\caption{BERT}
\label{fig:tsne_c}
\end{subfigure}~
\begin{subfigure}{0.24\textwidth}
\includegraphics[width=\textwidth]{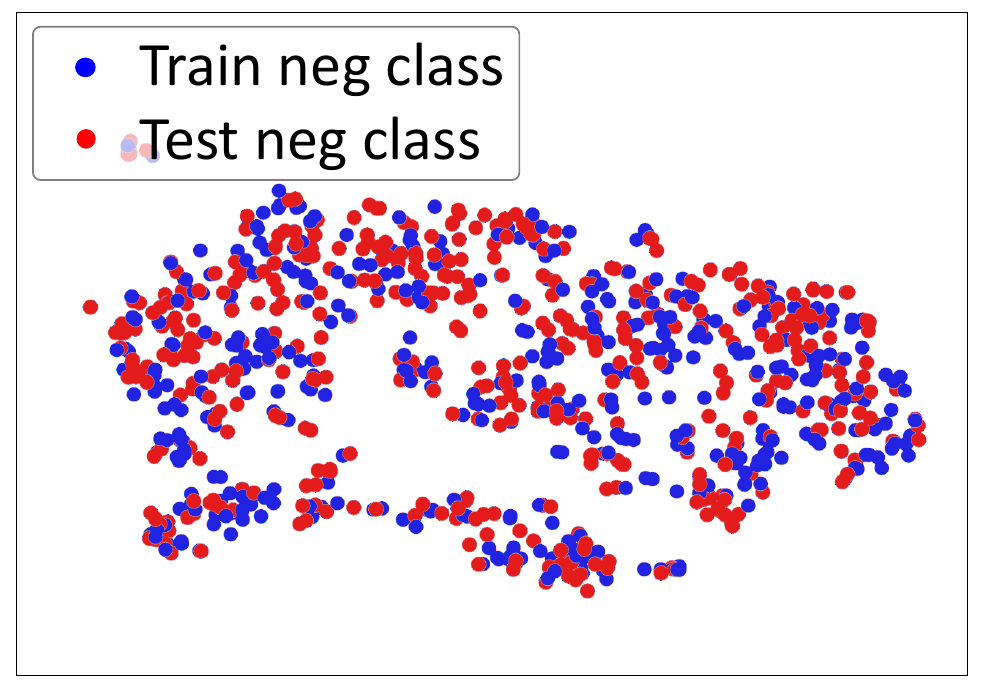}
\caption{MASKER (ours)}
\label{fig:tsne_d}
\end{subfigure}
\caption{
t-SNE plots on the document embeddings of BERT and MASKER, on (a,b) OOD detection (Amazon 50 class reviews with split ratio 25\%), and (c,d) cross-domain generalization (Fine Food to SST-2). (a,b) Blue and red dots indicate the in- and out-of-distribution samples, respectively. (c,d) Blue and red dots indicate the samples from the same classes (`negative') from training and test domains, respectively. MASKER better distinguishes OOD samples and entangles cross-domain samples.
}\label{fig:tsne}
\vspace{-0.05in}
\end{figure*}

In Figure \ref{fig:tsne_a} and Figure \ref{fig:tsne_b}, we visualize the t-SNE \citep{maaten2008visualizing} plots on the document embeddings of BERT and MASKER, under the Amazon reviews dataset with a split ratio of 25\%. Blue and red points indicate in- and out-of-distribution samples, respectively. Unlike the samples that are entangled in the vanilla BERT, MASKER clearly distinguishes the OOD samples.

\subsection{Cross-domain Generalization}
\label{sec:exp_gen}

We conduct the experiments on sentiment analysis (IMDB \citep{maas2011learning}; SST-2 \citep{socher2013recursive}; Fine Food \citep{mcauley2013amateurs}), natural language inference (MNLI, \citet{williams2017broad}), and semantic textual similarity (STS-B, \citet{wang2019glue} dataset) tasks, following the settings of \citet{hendrycks2020pretrained}.

Table~\ref{tab:gen_ablation} shows an ablation on MASKER study under the sentiment analysis task. The results are consistent with OOD detection, \eg, all components contribute to cross-domain generalization. Notably, while MER is not helpful for the original domain accuracy (see Table~\ref{tab:ood_ablation}), it improves the cross-domain accuracy for most settings. In particular, MASKER improves the cross-domain accuracy from 75.6\% to 80.0\% for Fine Food to SST-2 task. We analyze the most influential keywords (see Appendix~\ref{app:analysis}) and find that MASKER extracts the sentiment-related (\ie, generalizable) keywords (\eg, `astonishing') while the vanilla BERT is biased to some domain-specific words (\eg, `moonlight').

Table \ref{tab:gen_main} presents the results on sentiment analysis, natural language inference, and semantic textual similarity tasks. We compare MASKER with the vanilla fine-tuning and residual ensemble. The residual ensemble helps cross-domain generalization, but the gain is not significant and often degrades the original domain accuracy. This is because the keywords can be useful features for classification. Hence, nai\"vely removing (or debiasing) those features may lose the information. In contrast, MASKER facilitates contextual information rather than removing the keyword information, which regularizes the over-reliance in a softer manner.

In Figure \ref{fig:tsne_c} and Figure \ref{fig:tsne_d}, we provide the t-SNE plots on the document embeddings of BERT and MASKER, under the Fine Food to STS-2 task. Blue and red points indicate original and cross-domain samples, respectively. MASKER better entangles the same classes in training and test datasets (of the different domains) while BERT fails to do so.

\section{Conclusion}

The reliability of text classifiers is an essential but under-explored problem. We found that the over-reliance on some keywords can be problematic for out-of-distribution detection and generalization. We propose a simple yet effective fine-tuning method, coined masked keyword regularization (MASKER), composed of two regularizers and keyword selection schemes to address this issue. We demonstrate the effectiveness of MASKER under various scenarios.

\clearpage
\section*{Acknowledgements}

This work was supported by Center for Applied Research in Artificial Intelligence(CARAI) grant funded by Defense Acquisition Program Administration(DAPA) and Agency for Defense Development(ADD) (UD190031RD). 

\nocite{tack2020csi}
\bibliography{ref}

\clearpage
\appendix

\section{Experimental Details}
\label{app:detail}

\subsection{Training Details}

Following \citet{devlin2018bert,liu2019roberta}, we select the best hyperparameters from the search space below. We choose learning rate from \{$1\mathrm{e}{-5}$, $2\mathrm{e}{-5}$, $5\mathrm{e}{-5}$\} and batch size from \{$16$,$32$\}. We halve the learning rate for the embedding layers of MASKER since the regularizer for fits to the classifier, and directly updating the embedding layers can be unstable. We also use the batch size of 4 for random word reconstruction due to the large vocabulary size. We use the Adam \citep{kingma2015adam} optimizer for all experiments. We train vanilla BERT and ALBERT for 3$\sim$4 epochs, and RoBERTa for 10 epochs following \citet{devlin2018bert} and \citet{liu2019roberta}, respectively. For MASKER, we train BERT+MASKER, ALBERT+MASKER for 6$\sim$8 epochs, and train RoBERTa+MASKER for 12 epochs. We remark that all the models are trained until convergence. Since MER cannot be directly applied to the regression tasks (\eg, STS-B), we only use MKR for such settings.

\subsection{Dataset Details}

We use the pre-defined train and test splits if they exists: for IMDB \citep{maas2011learning}, SST-2 \citep{socher2013recursive}, MNLI \citep{williams2017broad}, and STS-B \citep{wang2019glue} datasets. If pre-defined splits not exists, we randomly divide the dataset with 70:30 ratio, using them for train and test splits, respectively: for 20 Newsgroups \citep{lang1995newsweeder}, Amazon 50 class reviews \citep{chen2014mining}, and Fine Food \citep{mcauley2013amateurs} datasets. We do not use any pre-processing methods, \eg, removing headers.

\subsection{Evaluation Metrics}

Let TP, TN, FP, and FN denotes true positive, true negative, false positive, and false negative, respectively. We use the following metrics for OOD detection:
\begin{itemize}
    \item \textbf{Area under the receiver operating characteristic curve (AUROC).}
    The ROC curve is a graph plotting true positive rate (TPR) = TP / (TP+FN) against the false positive rate (FPR) = FP / (FP+TN) by varying a threshold. AUROC measures the area under the ROC curve.
    \item \textbf{Equal error rate (EER).} EER is the error rate when the confidence threshold is located where FPR is the same with the false negative rate (FNR) = FN / (TP+FN).
    \item \textbf{Detection accuracy.} Measures the maximum classification probability among all possible threshold sets.
    \item \textbf{TNR at TPR 80\%.} Measures true negative rate (TNR) = TN / (FP+TN) when TPR = 80\%.
\end{itemize}
AUROC measures the overall performance varying thresholds, and the other three metrics measure the performance for some fixed threshold.

\section{Overall Procedure of MASKER}
\label{app:overall}

\begin{figure*}[t]
\centering
\includegraphics[width=0.65\textwidth]{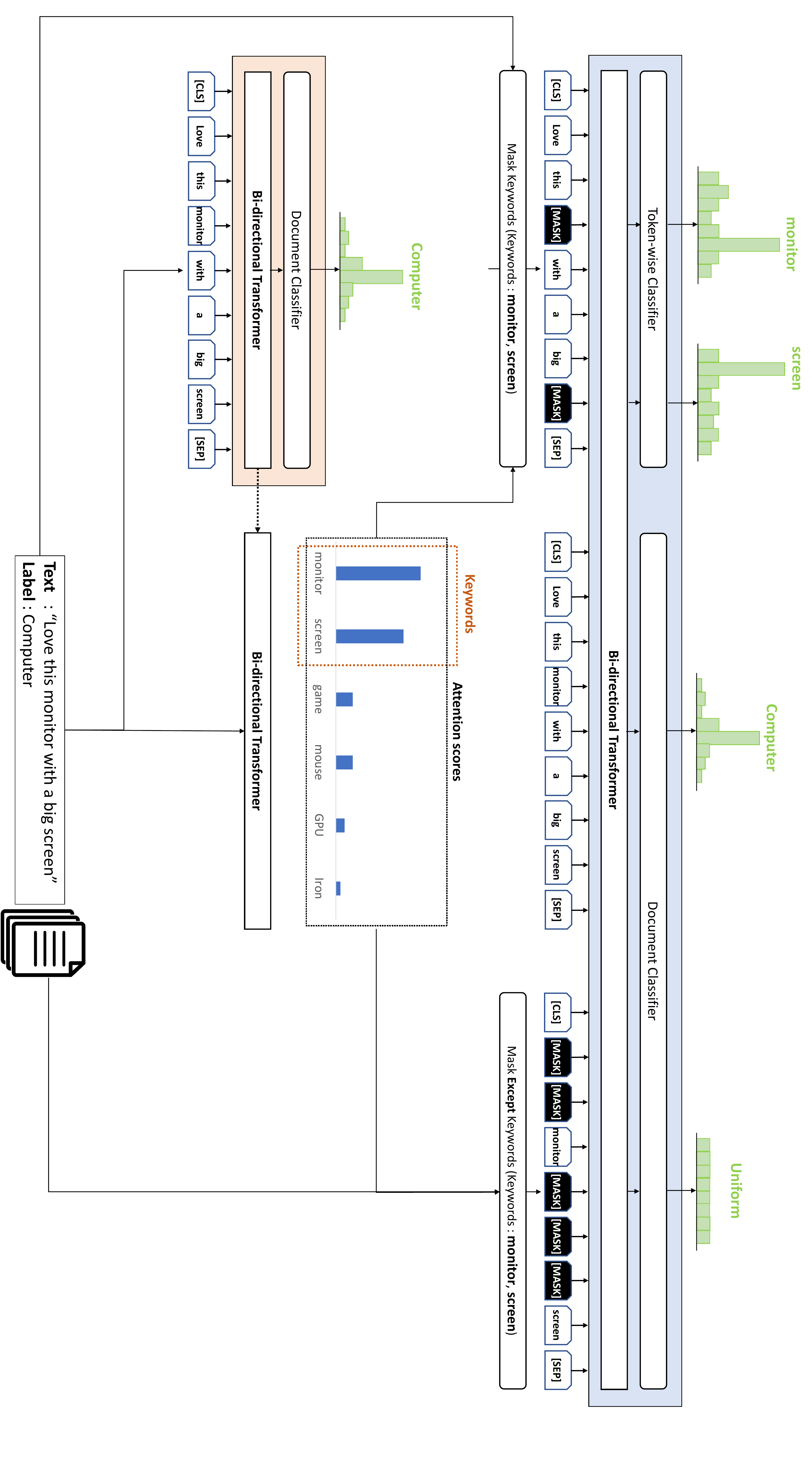}
\caption{
The overall procedure of MASKER using the attention-based keywords. Vanilla training and our proposed method are presented as the red and blue boxes, respectively. Parameter sharing without back-propagation is presented as a dashed arrow.
}\label{fig:overall}
\end{figure*}

Figure \ref{fig:overall} visualizes the overall procedure of MASKER, including the keyword selection scheme (attention-based selection using a vanilla model), masked keyword reconstruction, and masked entropy regularization.

\section{Additional Experimental Results}
\label{app:more_exp}

Table~\ref{tab:ood_more} presents the additional OOD detection results, including the split settings of 20 Newsgroups and Amazon reviews datasets \citep{shu2017doc}. Our method consistently outperforms the baseline methods. Table~\ref{tab:ood_acc} presents the classification accuracy of the baselines and MASKER, which validates that MASKER does not degrade the accuracy. Table~\ref{tab:gen_stsb_all} presents the other cross-domain results under the STS-B dataset, shows the effectiveness of our method. We also try to regularize attention weights to be uniform directly, but it harms both classification accuracy and OOD detection performance for vanilla models.

\section{Analysis on Attention Scores}
\label{app:analysis}

\begin{figure}[htb]
\centering
\includegraphics[width=0.45\textwidth]{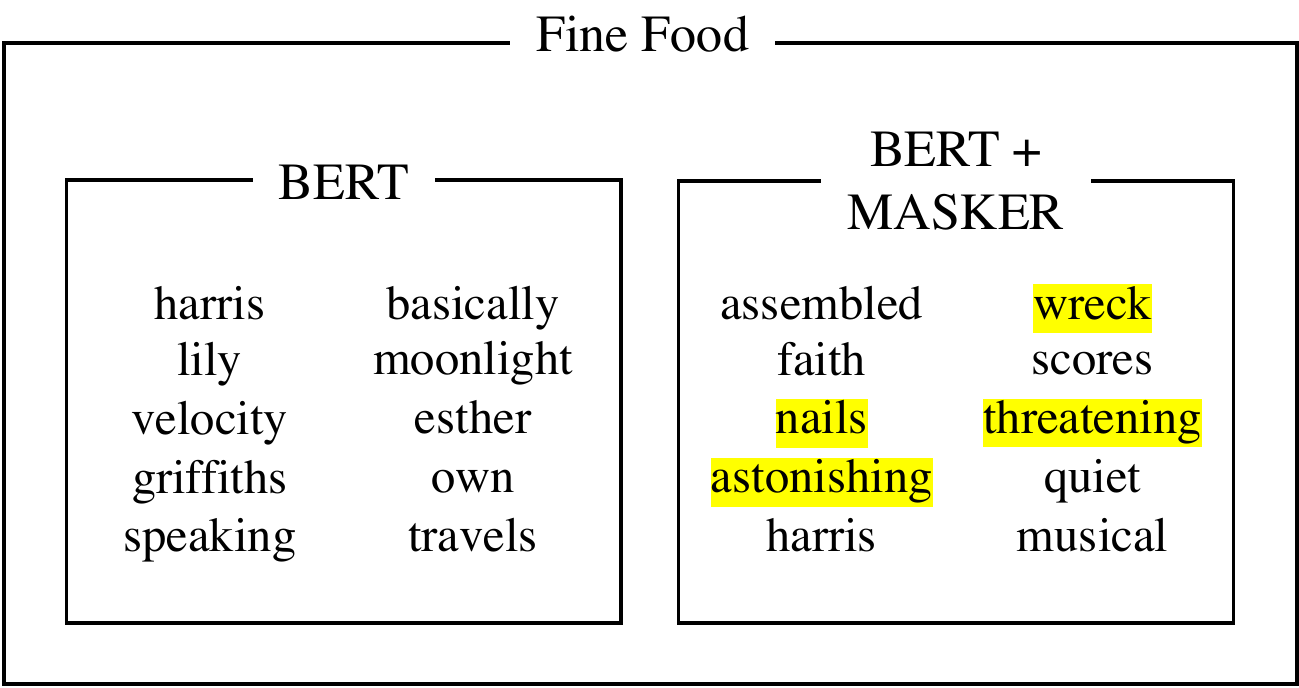}
\caption{
Top 10 keywords according to the attention scores, chosen by BERT and ours trained on Fine Food and tested on
SST-2. The sentiment-related keywords are highlighted.
}\label{fig:analysis}
\end{figure}

In Figure~\ref{fig:analysis}, we visualize the most influential keywords measured by the attention scores. Our method makes predictions based on the generalizable keywords, \eg, sentiment-related keywords for sentiment analysis tasks.

\clearpage

\clearpage
\begin{table*}[ht]
\centering\small
\begin{tabular}{ccccccc|c}
\toprule
ID  & OOD & Split Ratio & AUROC $\uparrow$ & EER $\downarrow$ & \makecell{Detection\\Accuracy $\uparrow$} & \makecell{TNR at\\ TPR 80\% $\uparrow$} & \makecell{Classification\\ Accuracy $\uparrow$} \\
\cmidrule{4-8}
& & & \multicolumn{5}{c}{DOC / BERT / BERT+MAKSER (ours)} \\
\midrule
\multirow{8}{*}{Newsgroup} & \multirow{3}{*}{Newsgroup}
    & 10\% & 83.7/85.4/\textbf{87.0} & 23.1/22.9/\textbf{21.0} & 81.0/90.9/\textbf{91.5} & 61.0/67.4/\textbf{75.2} & 98.7/99.0/98.9 \\
  & & 25\% & 86.1/89.0/\textbf{91.0} & 18.7/19.1/\textbf{17.0} & 82.7/83.0/\textbf{86.9} & 78.5/80.5/\textbf{84.8} & 93.9/95.3/94.9 \\
  & & 50\% & 80.4/82.4/\textbf{83.0} & 28.2/25.4/\textbf{25.0} & 72.3/75.0/\textbf{76.8} & 65.0/68.0/\textbf{71.8} & 89.9/94.4/94.0 \\
  \cmidrule(lr){2-8}
  & Amazon & \multirow{5}{*}{100\%}
             & 84.1/86.0/\textbf{96.8} & 23.3/19.5/\textbf{8.3} & 90.1/87.2/\textbf{94.7} & 74.5/86.7/\textbf{95.5} & \multirow{5}{*}{86.9/90.4/90.1} \\
  & Reuter & & 60.0/91.8/\textbf{97.7} & 41.1/14.7/\textbf{6.7} & 75.2/85.7/\textbf{93.6} & 21.3/84.5/\textbf{96.4} & \\
  & IMDB   & & 88.6/94.6/\textbf{98.5} & 19.1/11.5/\textbf{5.1} & 88.4/93.4/\textbf{96.3} & 81.7/87.7/\textbf{98.2} & \\
  & SST-2  & & 88.1/87.0/\textbf{98.6} & 18.7/18.9/\textbf{5.1} & 86.6/88.5/\textbf{96.0} & 81.8/92.4/\textbf{98.4} & \\
  & Fine Food  & & 81.3/85.3/\textbf{93.4} & 25.7/19.8/\textbf{10.9} & 74.8/82.7/\textbf{90.5} & 67.6/85.9/\textbf{95.2} \\
\midrule
\multirow{8}{*}{Amazon} & \multirow{3}{*}{Amazon}
    & 10\% & 80.7/82.8/\textbf{86.0} & 22.9/22.1/\textbf{21.0} & 79.2/92.3/\textbf{93.0} & 74.0/75.7/\textbf{80.1} & 92.7/93.5/93.3 \\
  & & 25\% & 75.1/78.5/\textbf{85.5} & 25.6/28.3/\textbf{22.3} & 80.2/81.2/\textbf{85.0} & 66.0/66.3/\textbf{74.1} & 83.1/84.8/85.2 \\
  & & 50\% & 74.6/75.4/\textbf{77.7} & 29.0/28.9/\textbf{27.9} & 69.4/70.2/\textbf{71.1} & 60.0/60.7/\textbf{61.5} & 78.8/81.7/80.6 \\
  \cmidrule(lr){2-8}
  & Newsgroup & \multirow{5}{*}{100\%}
              & 81.3/84.8/\textbf{87.2} & 23.0/21.6/\textbf{20.0} & 80.0/81.5/\textbf{83.5} & 72.7/77.0/\textbf{80.4} & \multirow{5}{*}{66.6/70.8/70.0}\\
  & Reuter    & & 79.8/89.7/\textbf{93.5} & 30.8/18.1/\textbf{12.2} & 82.5/87.1/\textbf{89.9} & 60.0/83.2/\textbf{90.7} \\
  & IMDB      & & 89.6/93.3/\textbf{95.2} & 18.1/13.9/\textbf{10.5} & 82.4/87.0/\textbf{90.7} & 83.0/89.2/\textbf{92.9} \\
  & SST-2 & & 91.5/93.0/\textbf{95.6} & 15.8/14.1/\textbf{9.5} & 84.5/89.6/\textbf{92.9} & 87.1/88.2/\textbf{93.9} \\
  & Fine Food  & & 66.8/78.5/\textbf{84.9} & 38.0/30.0/\textbf{19.5} & 69.7/73.9/\textbf{80.7} & 55.3/62.7/\textbf{80.8} \\
\bottomrule
\end{tabular}
\caption{
AUROC (\%) on various additional OOD detection scenarios. The reported results are averaged over three trials, and the best results are highlighted in bold. MASKER outperforms the baselines in all cases.
}\label{tab:ood_more}
\end{table*}

\begin{table*}[ht]
\centering\small
\begin{tabular}{c|cc|c|c|c}
\toprule
\multirow{2}{*}{Dataset} & \multirow{2}{*}{Openmax} & \multirow{2}{*}{DOC} & \multicolumn{3}{c}{Vanilla/Residual/MASKER} \\
& & & BERT & RoBERTa & ALBERT \\ \cmidrule{1-6}
Newsgroups & 85.8 & 86.9 & 90.4/89.5/90.1 & 90.9/88.1/90.7 & 89.3/89.7/89.7 \\
Amazon & 63.0 & 66.6 & 70.8/70.3/70.0 & 71.0/69.1/71.2 & 68.7/64.2/68.6 \\
\bottomrule
\end{tabular}
\caption{
Classification accuracy (\%) of the datasets used for OOD detection. MASKER shows a comparable accuracy with the vanilla models, while residual ensemble shows a marginal drop.
}\label{tab:ood_acc}
\end{table*}

\begin{table*}[ht]
\centering\small
\begin{tabular}{c|cccc|cccc}
\toprule
\multirow{2}{*}{Model} & \multicolumn{4}{c}{MSRvid} & \multicolumn{4}{c}{Images} \\
& MSRvid & Images & MSRpar & Headlines & Images & MSRvid & MSRpar & Headlines \\
\cmidrule(lr){1-9}
BERT & 91.5 & 82.0 & 38.2 & 61.7 & 88.0 & 89.7 & 50.8 & 73.9 \\
+MASKER & 91.2 & \textbf{84.3} & \textbf{40.9} & \textbf{66.7} & 88.1 & \textbf{91.6} & \textbf{52.5} & \textbf{75.3} \\
\cmidrule(lr){1-9}
RoBERTa & 94.2 & \textbf{88.0} & 66.0 & 80.3 & 91.8 & 92.9 & 68.4 & 84.1 \\
+MASKER & 93.7 & \textbf{88.0} & \textbf{67.1} & \textbf{84.0} & 91.3 & \textbf{94.1} & \textbf{70.1} & \textbf{85.3} \\
\cmidrule(lr){1-9}
ALBERT & 92.6 & 81.2 & 39.4 & 60.6 & 90.4 & 90.9 & 44.9 & 69.8 \\
+MASKER & 93.3 & \textbf{82.6} & \textbf{39.8} & \textbf{68.8} & 90.5 & \textbf{92.0} & \textbf{45.2} & \textbf{78.4} \\
\bottomrule
\toprule
\multirow{2}{*}{Model} & \multicolumn{4}{c}{Headlines} & \multicolumn{4}{c}{MSRpar} \\
& Headlines & MSRvid & Images & MSRpar & MSRpar & MSRvid & Images & Headlines \\
\cmidrule(lr){1-9}
BERT & 86.1 & 83.2 & 81.1 & 69.9 & 74.2 & 74.1 & 71.9 & 67.1 \\
+MASKER & 86.8 & \textbf{88.0} & \textbf{83.6} & \textbf{75.8} & 77.6 & \textbf{79.1} & \textbf{75.9} & \textbf{67.7}\\
\cmidrule(lr){1-9}
RoBERTa & 90.7 & \textbf{93.3} & 90.1 & \textbf{75.5} & 86.4 & 88.2 & 85.8 & 85.4 \\
+MASKER & 88.2 & 90.3 & \textbf{90.7} & 70.9 & 84.8 & \textbf{90.2} & \textbf{85.9} & \textbf{86.4} \\
\cmidrule(lr){1-9}
ALBERT & 86.8 & 89.3 & \textbf{87.1} & 63.5 & 78.5 & 82.4 & 80.8 & 69.2 \\
+MASKER & 87.0 & \textbf{90.4} & \textbf{87.1} & \textbf{67.5} & 76.7 & \textbf{82.7} & \textbf{81.7} & \textbf{75.8}\\
\bottomrule
\end{tabular}
\caption{
Total Pearson correlation (\%) of four genres in the STS-B dataset. The reported results are averaged over 3 trials.
}\label{tab:gen_stsb_all}
\end{table*}

\end{document}